\documentclass{article}
\usepackage[utf8]{inputenc}
\usepackage[T1]{fontenc}
\usepackage{lmodern}
\usepackage{authblk}
\usepackage{setspace}
\usepackage[margin=1in]{geometry}
\usepackage{graphicx}
\graphicspath{ {./figures/} }
\usepackage{subcaption}
\usepackage{amsmath}
\usepackage{amssymb}
\usepackage{url}
\usepackage{float}
\usepackage{booktabs}
\usepackage{tabularx}
\usepackage[font=small,labelfont=bf,labelsep=period]{caption}

\usepackage[style=nejm,
citestyle=numeric-comp,
sorting=none,
backend=bibtex]{biblatex}
\usepackage[hidelinks]{hyperref}

\title{Simulating Infant Senses with Age-Tuned Networks}

\author[1*]{Somesh Mohapatra}

\affil[1]{Independent.}
\affil[*]{Address correspondence to: pikulsomesh@gmail.com}

\date{}

\newcolumntype{L}{>{\raggedright\arraybackslash}X}

\begin{document}

\maketitle

\begin{abstract}
What a newborn sees, hears and feels, and how this changes month by month, has been measured sense by sense, with different methods and in different units. We bring these measurements together for the first two years of life and ask whether small neural networks tuned to a given age can turn them into a quick, inspectable simulation of sensory perception. Each sense's immaturity is written as a fixed, age-indexed filter on the input, and a small prediction network learns behind it over the first year. The filters alone reproduce the order in which the senses mature, by construction. The networks add what filters cannot: in vision they extract far less than their input allows at early stages (at the newborn stage a toy's position cannot be read out, $R^2 = 0.00$, against 0.57 for a network trained directly on the same input), and a critical period and perceptual narrowing emerge without being built in; in hearing they track what their input allows. The same filters drive the Baby Senses Lab, a free website that shows a user's own images and sounds as a baby of a chosen age might sense them. It is a research and teaching tool, not a diagnostic test.
\end{abstract}


\section{Introduction}

A newborn's senses are not small versions of an adult's. They differ in kind and in timing. Touch is already working before birth \autocite{bremner2017development}, and smell learns from the mother's diet: newborns prefer odours their mother ate late in pregnancy \autocite{schaal2000human}. Hearing begins in the womb, so that newborns already prefer their mother's voice to another woman's \autocite{decasper1980human}. Vision has had no patterned light before birth and starts far behind: a newborn resolves roughly one stripe pair per degree of visual angle, about thirty times coarser than an adult, and sees faint contrasts and many colours poorly \autocite{dobson1978visual,banks1988optical,adams1994systematic}. Over the first year each sense improves on its own clock, and then tunes itself to the faces, voices and speech sounds around it \autocite{werker1984cross,kuhl1992linguistic,pascalis2002species,maurer2014perceptual}.

This staggered development, sometimes called sensory heterochrony, is one of the best documented facts in developmental neuroscience, and it may matter for how the senses organise themselves: limited and staggered early input can reduce competition between senses \autocite{turkewitz1982limitations}, and redundancy across senses directs early attention \autocite{bahrick2000intersensory}. Yet the evidence is hard to use as a whole. Each sense is measured with its own method (preferential looking, evoked potentials, head turns, sucking rate, facial expressions), in its own units (cycles per degree, decibels, looking time), and over its own age range. A parent, teacher, clinician or modeller who wants to know \emph{what a three-month-old senses} has to assemble the answer from many sources, and still cannot see or hear it.

Two developments make it possible to do better. First, much of sensory immaturity at the periphery can be written down as a filter: optical and photoreceptor immaturity limit spatial and chromatic vision in a way that can be modelled directly \autocite{banks1988optical}, and hearing thresholds and frequency resolution can be expressed as attenuation and smearing \autocite{werner2007issues}. Second, neural networks trained on sensory data have become useful models of sensory systems \autocite{yamins2016using}, and networks trained on infant-like input reproduce some developmental effects: blurred early input changes what visual networks learn \autocite{vogelsang2018potential,vogelsang2024impact,avbersek2021training,jang2021convolutional}, networks show critical periods for deprivation and for combining senses \autocite{achille2019critical,kleinman2023critical}, and networks trained on head-camera video from young children learn useful visual and word representations \autocite{orhan2024learning,vong2024grounded,sheybani2023curriculum}. Networks that learn by forecasting their own sensory input are small instances of the predictive world models that machine learning now treats as a route to systems that understand their surroundings \autocite{ha2018world,lecun2022path,li2025words}; here they serve a narrower purpose, as models of what each sense can extract at a given age.

This paper joins the two. Its scope is \emph{sensory perception}: what information about the world reaches and is extracted by each sense at a given age. We do not model cognition (object permanence, goals, number, word learning), and where our models touch on it we say so. We first assemble a sense-by-sense synthesis of sensory milestones from birth to 24 months (Fig.~\ref{fig:timeline}, Table~\ref{tab:senses}), written so that each entry can be turned into a number a simulation can use and stating the measurement method, because different methods give different answers, most clearly for visual acuity \autocite{dobson1978visual}. We then pass each sense through a fixed, age-indexed front end (Fig.~\ref{fig:frontends}) and train a small prediction network behind it, over the first 12 months in the main models and to 24 months in a follow-up, and show what the front ends alone explain, what the networks add and where the two come apart (Figs.~\ref{fig:ceiling} and \ref{fig:experience}). Finally, we describe the Baby Senses Lab, a free, browser-only website that uses the same front ends to show a user's own images, videos and sounds as a baby of a chosen age might sense them, together with age profiles, an activity library and short observation sessions (Fig.~\ref{fig:app}); it is meant for exploration and research and does not screen or diagnose.

\noindent\textbf{Relation to the companion paper.} This paper and a companion paper \autocite{mohapatra2026learn} come from the same project and are written for different questions and readers. The companion paper is about machine learning: how predictive world models learn under developmental input constraints, with its own analyses of learning dynamics. This paper is about developmental neuroscience and sensory perception. What is new here is the sense-by-sense synthesis (Fig.~\ref{fig:timeline}, Table~\ref{tab:senses}), the comparison of behavioural and evoked-potential acuity and the extension of the front ends to 24 months and to real images and sound (Fig.~\ref{fig:frontends}), the per-sense comparison of available and learned information (Fig.~\ref{fig:ceiling}), the cross-sense and age-modelling results of Fig.~\ref{fig:experience}, and the Baby Senses Lab (Fig.~\ref{fig:app}). What is shared is stated where it occurs: the vision front-end values and equations, and the main-study training runs and information ceilings behind Fig.~\ref{fig:ceiling} and the maturation-order measure. No figure is reproduced from the companion paper.

\section{Results}

\subsection{Each sense starts at a different point and improves on its own clock}

\begin{figure}[tbp]
\centering
\includegraphics[width=\textwidth]{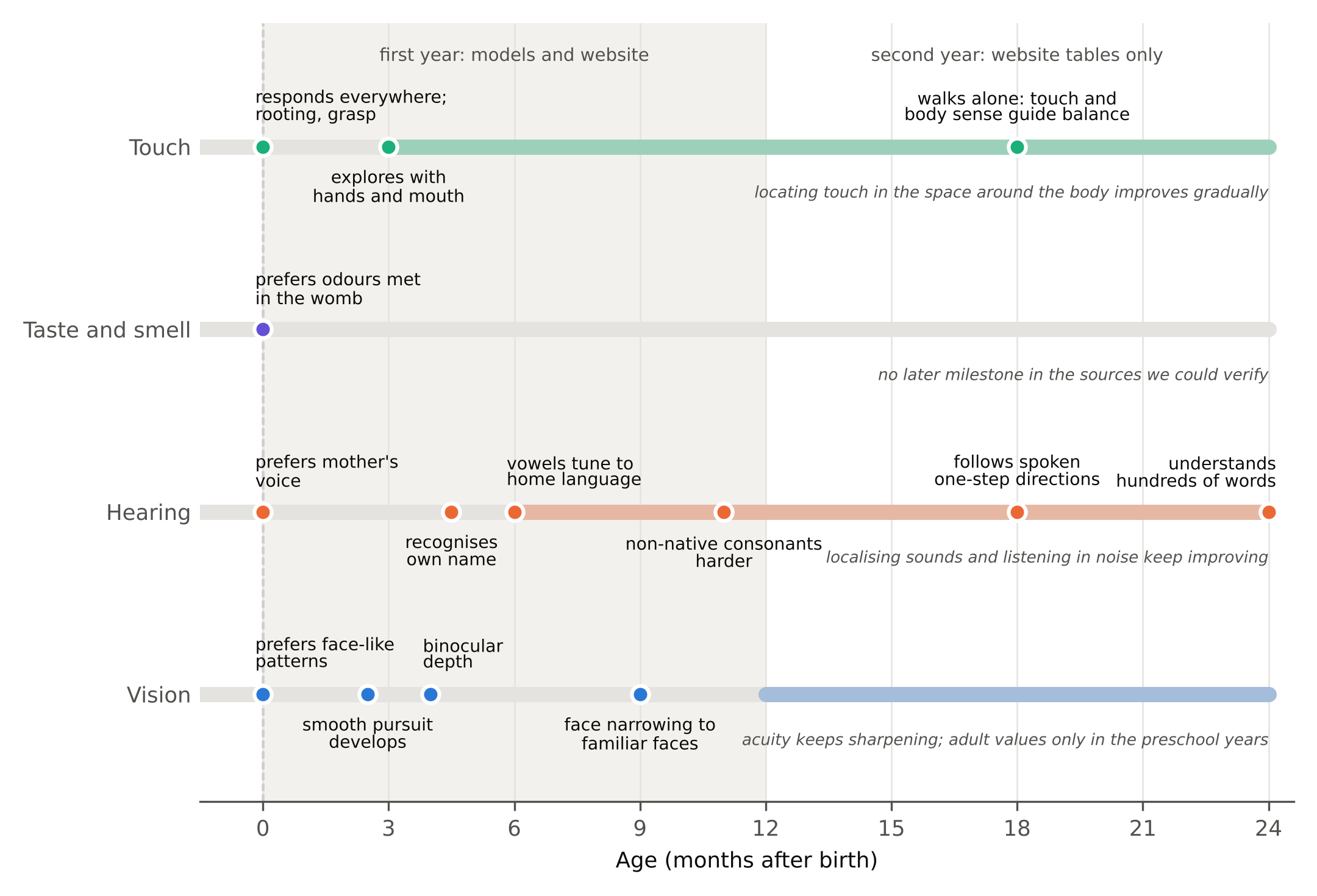}
\caption{\textbf{Sensory milestones from birth to 24 months, by sense.} Each dot is a typical age at which a capacity is reported in group studies; individual babies vary by weeks to months. Touch: responsiveness and reflexes at birth, exploration with hands and mouth, then with fingers \autocite{bremner2017development}. Taste and smell: preference at birth for odours of the mother's diet in late pregnancy \autocite{schaal2000human}. Hearing: preference for the mother's voice at birth \autocite{decasper1980human}, recognition of own name at about 4.5 months \autocite{mandel1995infants}, vowel perception shaped by the home language by 6 months \autocite{kuhl1992linguistic}, decline of non-native consonant discrimination by 10--12 months \autocite{werker1984cross}. Vision: preference for face-like patterns at birth \autocite{fantz1963pattern,johnson1991newborns}, smooth pursuit developing between 2 and 3 months \autocite{vonhofsten1997development} and colour discrimination improving over the first months \autocite{adams1994systematic,braddick2011development}, binocular depth (stereopsis) at about 4 months \autocite{held1980stereoacuity}, face perception narrowing to familiar species and groups by 9 months \autocite{pascalis2002species,kelly2007other}. Second year: walking alone, with touch and body sense guiding balance, and following one-step spoken directions by about 18 months \autocite{zubler2022evidence}; comprehension of hundreds of words by 24 months \autocite{frank2021variability,zubler2022evidence}. Horizontal bars mark gradual refinement without a single milestone age: locating touch in space \autocite{bremner2017development}, sound localisation and listening in noise \autocite{werner2007issues}, and visual acuity \autocite{mayer1995monocular,teller1997first}. No later taste or smell milestone was found in the sources we could verify. Shading marks the first year, the range of the main networks; the second year is covered by the website's tables and by one follow-up run (Fig.~\ref{fig:beyond}a).}
\label{fig:timeline}
\end{figure}

Figure~\ref{fig:timeline} places the best documented sensory milestones of the first two years on one axis, and Table~\ref{tab:senses} gives the numbers behind them. The order is the one described for birds and mammals generally: touch and the chemical senses first, hearing next, vision last \autocite{turkewitz1982limitations}. Most discrete sensory milestones fall in the first year. In the second year the change is mostly gradual refinement: acuity keeps rising towards adult values, which are reached only in the preschool years \autocite{mayer1995monocular,teller1997first}; localising sounds and hearing in noise keep improving \autocite{werner2007issues}; and touch is increasingly located in the space around the body \autocite{bremner2017development}. The second-year milestones that are reported tend to combine a sense with action or language, such as walking alone and following spoken directions \autocite{zubler2022evidence}. The main networks stop at the 12-month stage (shaded in Fig.~\ref{fig:timeline}); a follow-up run that continues to 24 months is reported in the Discussion (Fig.~\ref{fig:beyond}a).

\paragraph{Touch.} Touch is among the first senses to function and is well developed at birth, when newborns respond to touch and show rooting and grasp reflexes. Over the first months touch becomes a means of exploration with the mouth and hands, and many tactile abilities, such as locating a touch in the space around the body, develop gradually through infancy in interaction with vision and hearing \autocite{bremner2017development}. Because touch is well developed at birth, it is the sense least changed by a developmental simulation, and we model it with only mild blur and noise.

\paragraph{Taste and smell.} The chemical senses also work before birth. Odours of the mother's diet reach the fetus, and newborns whose mothers ate anise in late pregnancy prefer its odour at birth, whereas other newborns avoid it or are neutral \autocite{schaal2000human}. These senses are not modelled in this work or in the website, which we state on every page that discusses senses; we include them here because a sense-by-sense account that omits them gives a misleading picture of what a newborn perceives.

\paragraph{Hearing.} Hearing begins in the womb, where sound reaches the fetus filtered by the mother's body, so that rhythm and melody pass better than fine phonetic detail. Newborns prefer their mother's voice to another woman's \autocite{decasper1980human}, which shows that prenatal listening has already shaped perception. Hearing is far from adult, however. In the first months infants need sounds to be much louder than adults do (at 1 month, about 35 dB more at 4 kHz); many basic abilities, including frequency resolution, are roughly adult-like by about 6 months, while sound localisation and listening in noise take longer \autocite{werner2007issues}. Experience then tunes speech perception: by 6 months vowel perception reflects the home language \autocite{kuhl1992linguistic}, and by 10--12 months discrimination of many non-native consonant contrasts declines \autocite{werker1984cross}. By about 4.5 months infants recognise the sound pattern of their own name \autocite{mandel1995infants}.

\paragraph{Vision.} Vision is the least mature sense at birth and changes the most in the first year. Newborn acuity is roughly 1 cycle per degree (about 20/600 in Snellen terms) and contrast sensitivity is low, in large part because of optical and photoreceptor immaturity \autocite{banks1988optical}. Two families of measurement disagree on how fast acuity then improves. Behavioural methods, in which an observer judges which of two patterns a baby prefers to look at, give values that rise from about 1 cycle per degree at 1 month to a few cycles per degree at 6 months, with adult values only reached in the preschool years \autocite{mayer1995monocular,teller1997first}. Visual evoked potentials, recorded from the scalp, give higher values at every age and a much steeper rise in the first year \autocite{norcia1985spatial,dobson1978visual}. Colour discrimination is limited at birth \autocite{adams1994systematic} and improves markedly over the first months \autocite{braddick2011development}. Smooth tracking of a slowly moving target develops mainly between 2 and 3 months \autocite{vonhofsten1997development}, and binocular depth from stereopsis emerges at around 4 months \autocite{held1980stereoacuity}. From birth, babies look preferentially at patterned and face-like stimuli \autocite{fantz1963pattern,johnson1991newborns}; over the first year face perception narrows toward the kinds of faces a baby sees \autocite{pascalis2002species,kelly2007other}. Visual development depends on visual input: in infants treated for dense congenital cataracts, acuity begins to improve within an hour of first patterned input \autocite{maurer1999rapid}.

\paragraph{Combining the senses.} Infants detect information that is the same across senses (rhythm, tempo, synchrony) before they detect many sense-specific properties, which guides early attention \autocite{bahrick2000intersensory}. Multisensory perception also narrows with experience over the first year \autocite{lewkowicz2009emergence}, in parallel with narrowing within speech and faces \autocite{maurer2014perceptual}.

\paragraph{The brain behind the senses.} These timetables reflect both the sense organs and the brain. Synapses are overproduced and then pruned, and the peak is reached earlier in auditory than in prefrontal cortex \autocite{huttenlocher1997regional}. Sensitive periods open and close at different times for different senses and levels of processing \autocite{hensch2005critical,werker2015critical}. On the interactive-specialisation view of functional brain development, regions begin broadly tuned and specialise through experience and competition \autocite{johnson2001functional}, which is the kind of process a learning network can simulate.

\begin{table}[t]
\caption{\textbf{What each sense can do, by age.} Values are typical for groups of babies; individual babies vary widely. For vision, two numbers are given where behavioural (preferential looking) and evoked-potential measures differ. Sources are given in the text; the numbers used by the simulations are in Fig.~\ref{fig:frontends}.}
\label{tab:senses}
\centering
\footnotesize
\begin{tabularx}{\textwidth}{@{}p{1.55cm}LLLL@{}}
\toprule
Sense & Birth & 3 months & 6 months & 12--24 months \\
\midrule
Vision: detail & $\approx$1 cycle/degree (20/600); sees bold shapes at 20--30 cm & 2.6 (looking) to 4 (evoked) cycles/degree & 4.8 (looking) to 15 (evoked) cycles/degree & 7.5--10.5 (looking); evoked responses near adult; fine detail improves into preschool \\
Vision: contrast, colour & only strong contrasts; colour limited & medium contrasts; colour much improved & most everyday contrasts; colour near adult & contrast near adult \\
Vision: motion, depth, faces & notices motion; prefers face-like patterns & smooth pursuit; stereopsis begins (3--5 months) & adult-like motion; uses depth when reaching & face perception narrowed to familiar faces (9 months) \\
Hearing & hears from before birth; quiet sounds missed; prefers mother's voice & thresholds still well above adult & many basic abilities near adult; vowels tuned to home language & non-native consonants harder (10--12 months); localisation and hearing in noise keep improving \\
Touch and body sense & most mature sense; responds everywhere; rooting and grasp & explores with hands and mouth & explores textures with fingers; combines body sense and vision in reaching & fine touch keeps sharpening; body sense supports standing and walking \\
Taste and smell & prefers odours met in the womb & -- & -- & -- (not modelled) \\
\bottomrule
\end{tabularx}
\end{table}

\subsection{Sensory immaturity can be written as an age-indexed filter}

\begin{figure}[tbp]
\centering
\includegraphics[width=\textwidth]{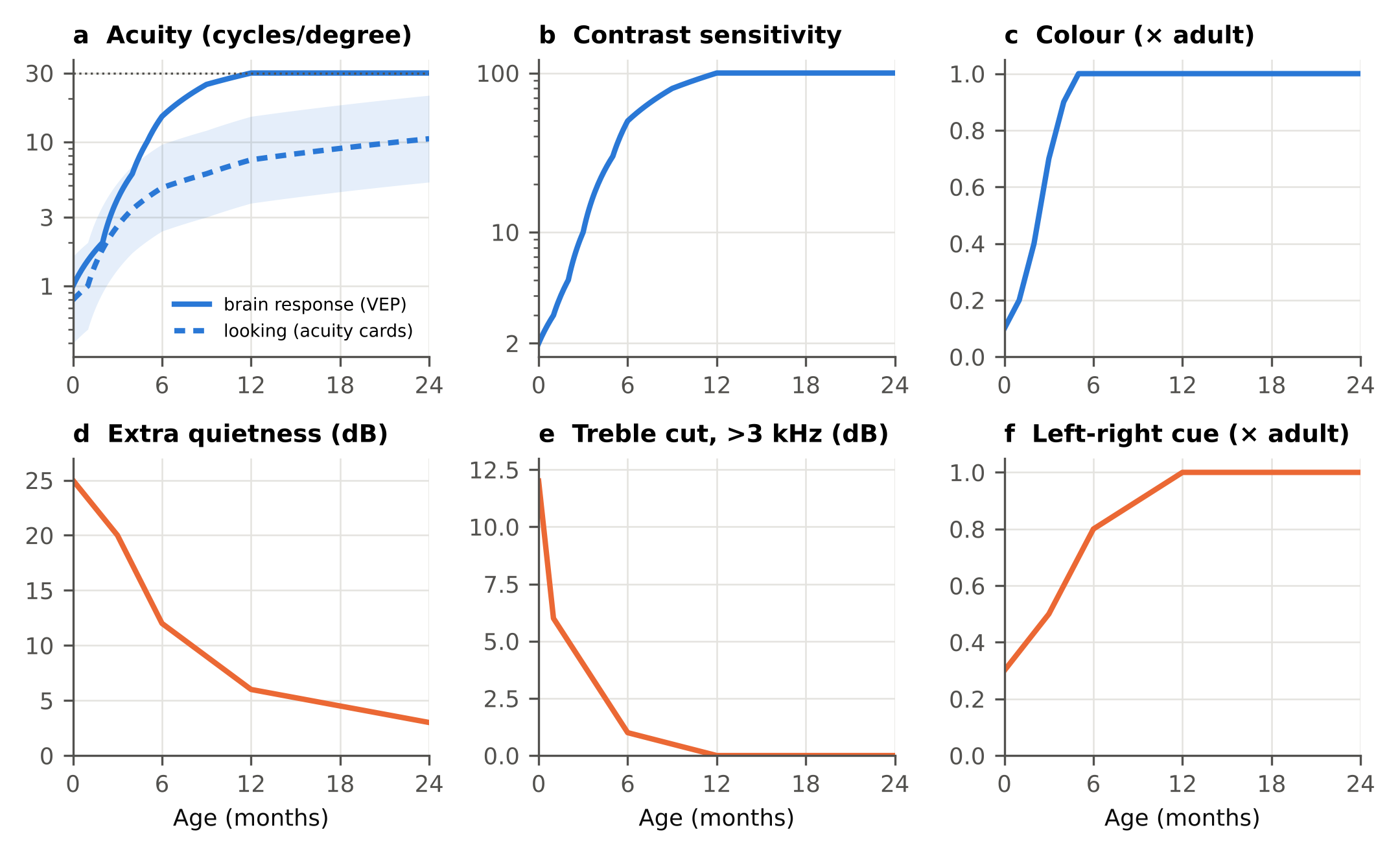}
\caption{\textbf{Age-tuned front ends used by the simulations and the website.} \textbf{a}, Visual acuity. Solid line: the evoked-potential-style table used by the models, which rises steeply in the first year \autocite{norcia1985spatial}; dashed line and band: behavioural acuity-card values, shown with the roughly one-octave normal range on either side \autocite{mayer1995monocular,teller1997first}; dotted line, 30 cycles/degree (adult reference). \textbf{b}, Contrast sensitivity. \textbf{c}, Colour saturation relative to adult (a simplification of chromatic sensitivity). \textbf{d}--\textbf{f}, Hearing: extra attenuation standing for raised thresholds \autocite{werner2007issues}, reduced high frequencies, and the fraction of the adult left-right (interaural) difference. Values between the listed ages are interpolated linearly. The vision tables follow published infant data; the hearing tables are coarse anchors that respect the direction of known changes, not fitted values; Fig.~\ref{fig:hearfit} compares them with threshold data. The evoked-potential acuity, contrast and colour values at 0, 3, 6 and 12 months are the same as the vision front-end table of the companion paper \autocite{mohapatra2026learn}; the behavioural acuity curve, the extension to 24 months and the hearing parameters used on real audio (\textbf{d}--\textbf{f}) are new here. Values beyond 12 months are used by the website and by the follow-up runs of Fig.~\ref{fig:beyond}a; the evoked-potential acuity, contrast and colour tables are held constant after 12 months (colour after 5), and only behavioural acuity and hearing change in the second year.}
\label{fig:frontends}
\end{figure}

To simulate a sense at a given age, we separate what is \emph{available} to the sense from what is \emph{learned}. Availability is set by a fixed front end $\Phi_s(\cdot\,;a)$ for each sense $s$ and age $a$ in months, which degrades an adult-quality input by the amounts reported for that age; learning is done by a network behind it. Figure~\ref{fig:frontends} shows the parameters shared by the models and the website.

For vision, the front end blurs, reduces contrast and desaturates (Materials and Methods, Eqs.~\ref{eq:blur}--\ref{eq:colour}). The blur is chosen so that detail finer than the age's acuity limit is removed: a Gaussian whose modulation transfer falls to 10\% at the acuity cut-off. Because acuity is an angle, how much a picture is blurred depends on how far away it is: at 30 cm, a 2-month-old's limit of about 2 cycles per degree removes stripes narrower than about 1.3 mm, while across a 2.5 m room it removes everything narrower than about 1 cm. This is why the website asks for viewing distance, and why its default view is the 20--30 cm at which young babies are usually held.

The two vision tables in Fig.~\ref{fig:frontends}a are a real ambiguity, not a modelling choice. At 6 months they differ by a factor of three (4.8 against 15 cycles per degree), because they measure different things: whether the brain responds to a pattern, and whether the baby's behaviour shows it was seen \autocite{dobson1978visual}. The models use the steeper table; the website shows both and lets the user switch, so that the simulation does not hide a disagreement in the data.

For hearing, the front end lowers the level, smooths and cuts high frequencies, narrows the difference between the ears and adds a noise floor. In the models these act on a 32-band binaural spectrum; in the website they act on real audio. Touch and body position receive blur and noise only, in line with their maturity at birth.

\subsection{The filters alone set the order in which the senses mature}

When the same prediction network is trained behind a newborn and a 12-month front end, the infant order of the senses appears without any change to the network (Fig.~\ref{fig:experience}a). Touch is at its 12-month level from the start (hidden-object tracking $R^2$ 0.38 at birth against 0.32 at 12 months), hearing distinguishes sound identity perfectly at both ages, and vision cannot track a hidden toy at all at birth ($R^2=-2.23$, worse than guessing the average position) but can at 12 months (0.33); three seeds per cell. Touch looks mature here because tracking a toy needs little tactile detail, so this is maturity for a coarse task and not a claim about fine touch. This order comes from the tables we give the front ends; it says nothing about learning. Its value is to show that per-sense input maturation is sufficient to produce the cross-sense ordering: no age-specific architecture is needed for it.

How the order is measured matters. If, instead of the learned networks, one asks at what age the information available in each sense's input reaches half of its 12-month value (an ``information ceiling'', estimated with a supervised network that is given the same filtered input), the infant order is \emph{not} recovered: touch and vision both reach half their final value at 0 months and hearing at 4 months (Kendall's $\tau = 0.00$ against touch before hearing before vision), because newborn visual input, although blurred, still contains enough coarse information to locate a large toy \autocite{mohapatra2026learn}. A sense can be immature for fine detail and adequate for coarse tasks at the same time; simulations should state which task defines ``mature''.

\subsection{Networks learn less than their visual input allows, and about as much as their auditory input allows}

\begin{figure}[tbp]
\centering
\includegraphics[width=\textwidth]{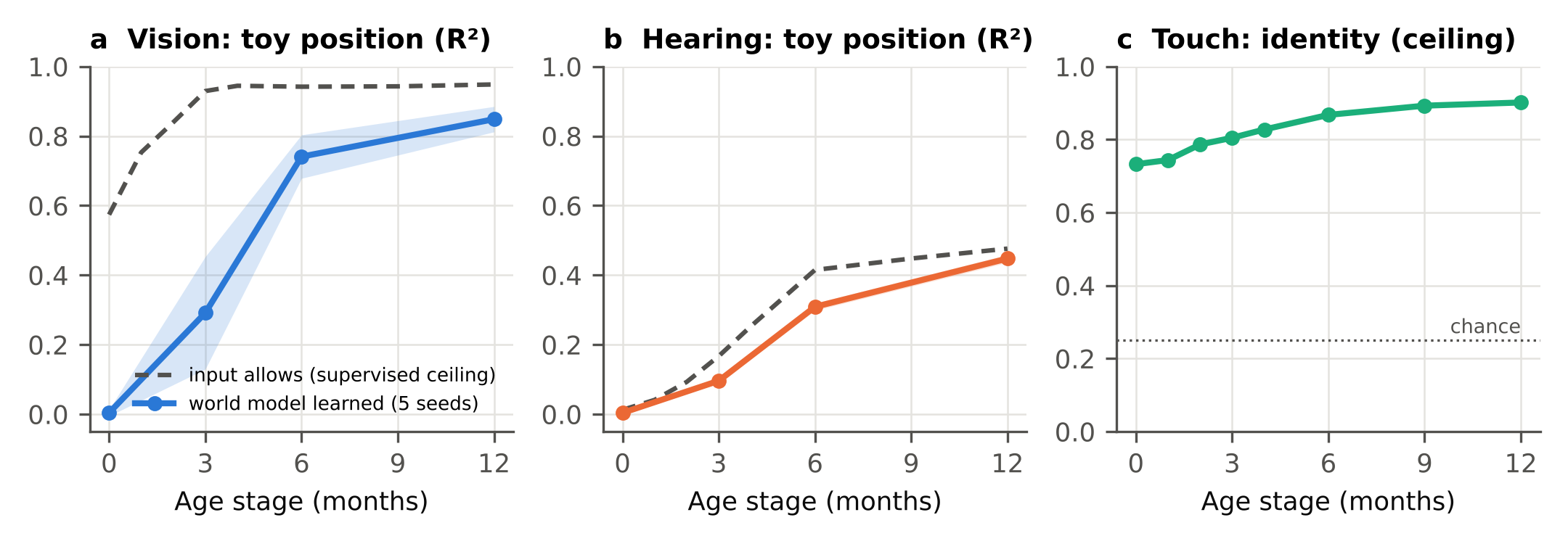}
\caption{\textbf{What the input allows and what a network learned, by age.} \textbf{a}, Vision: toy position decoded from a self-supervised prediction network trained through four age stages (solid; mean $\pm$ 1 s.d., 5 seeds; probe fitted and tested on input of the stage's own age), against a supervised network given only that age's filtered frames (dashed; one network per age). \textbf{b}, The same for hearing. \textbf{c}, Touch: toy identity from the supervised ceiling, high from birth (chance 0.25). Same training runs as the main study of the companion paper \autocite{mohapatra2026learn}, which reports these vision and hearing values in its development figure; this figure is a new plot that sets the learned and available values side by side for each sense.}
\label{fig:ceiling}
\end{figure}

Separating the front end from the learner lets us ask a question that infant studies cannot ask directly: is a young sense limited by its input or by what has been learned from it? Figure~\ref{fig:ceiling} compares, at each age stage, how well a toy's position can be decoded from what a predictive network has learned (a linear probe on its internal state, tested on input of the same age) with how well a network trained directly on the task does from the same filtered input.

The two senses behave differently. In vision, the newborn-stage network recovers essentially nothing about where the toy is ($R^2 = 0.00 \pm 0.01$), although the same blurred input supports 0.57 when a network is trained to find the toy; the gap is still large at 3 months ($0.29 \pm 0.16$ against 0.93) and closes by 12 months ($0.85 \pm 0.04$ against 0.95). In hearing, the learned network stays close to its ceiling at every stage ($0.10 \pm 0.01$ against 0.17 at 3 months, $0.45 \pm 0.01$ against 0.48 at 12 months). Early visual perception in these simulations is therefore limited twice: by the eye, and by how slowly a generic prediction objective picks out a small toy when the picture is blurred and faint. This distinction is relevant to infant data, where a low behavioural acuity can reflect either limit, and it is one reason the two acuity tables in Fig.~\ref{fig:frontends}a differ.

\subsection{Experience-dependent signatures of perception emerge from plain prediction}

\begin{figure}[tbp]
\centering
\includegraphics[width=\textwidth]{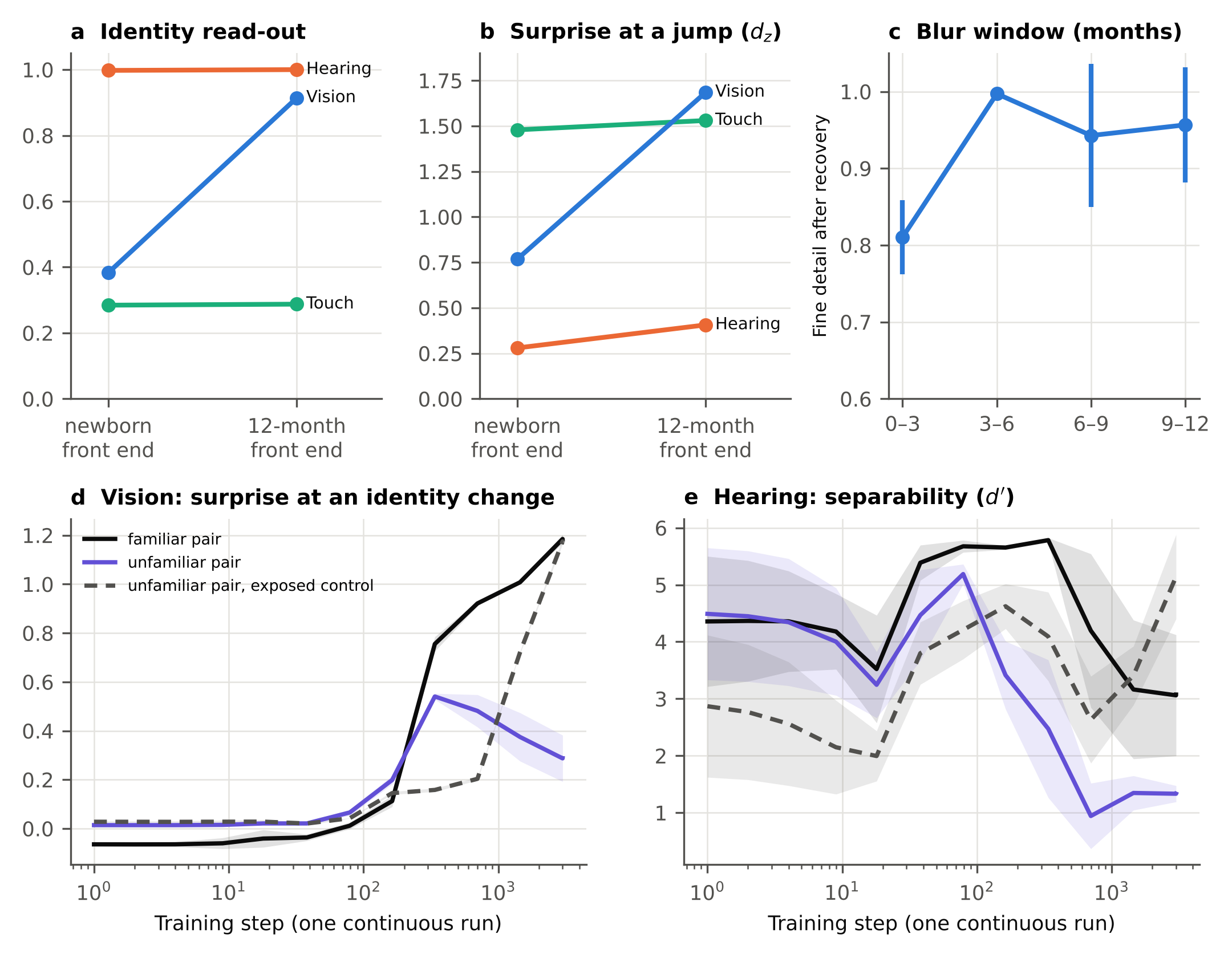}
\caption{\textbf{Perceptual signatures in age-tuned networks.} \textbf{a}, \textbf{b}, One recurrent prediction network trained behind a newborn and a 12-month front end in each sense: how well object identity can be read out (\textbf{a}) and how surprised the network is when an object jumps (\textbf{b}, $d_z$). Vision changes most; touch hardly changes. \textbf{c}, A critical period: final fine-detail accuracy after a 3-month window of heavy blur placed at different ages, each window followed by 8 months of clear input; only the earliest window leaves a lasting loss. \textbf{d}, \textbf{e}, Perceptual narrowing in a run where only two of four identities ever appear. In vision (\textbf{d}), surprise at a change between the two never-seen identities (violet) rises and then falls, while the familiar pair (black) keeps rising and a control that saw all identities (dashed) does not fall. In hearing (\textbf{e}) the same pattern appears in how separable the unfamiliar identities are ($d'$), not in surprise. Mean over 3 seeds (\textbf{a}--\textbf{c}) with $\pm$1 s.d. bars or bands. Drawn for this paper from the raw results of the project's cross-sense and age-modelling studies; this figure does not appear in the companion paper \autocite{mohapatra2026learn}.}
\label{fig:experience}
\end{figure}

Three further signatures of infant perception appear in the simulations without any age-specific mechanism (Fig.~\ref{fig:experience}). We report them as properties of small synthetic models that point to testable hypotheses, not as accounts of infant brains.

\paragraph{A critical period for visual detail.} In a vision network trained on a shape-versus-texture task, a 3-month window of heavy blur at the start of training reduced later fine-detail accuracy to $0.81 \pm 0.04$, whereas the same window at 3--6, 6--9 or 9--12 months was largely recovered from ($1.00 \pm 0.00$, $0.94 \pm 0.08$ and $0.96 \pm 0.06$; Fig.~\ref{fig:experience}c). Longer early deprivation produced larger, lasting losses (0.61 after 6 months, 0.40 after 9 months, with networks that stayed committed to coarse shape). This is the computational counterpart of lasting deficits after early visual deprivation and of critical periods in deep networks \autocite{achille2019critical}; it arises in one continuous training run with no plasticity schedule, and it cannot be shown by a set of separate models for each age, because no such model carries its history.

\paragraph{Perceptual narrowing.} When a network only ever sees two of four object identities, its surprise at a change between the two unseen identities first rises, as general features transfer, and then falls, as the representation specialises on the familiar ones (vision: from a peak of $0.54 \pm 0.01$ to $0.29 \pm 0.09$), while surprise at a change between familiar identities keeps rising ($1.18 \pm 0.02$) and a control that saw all four reaches the same level for the other pair ($1.17 \pm 0.01$; Fig.~\ref{fig:experience}d, e). This rise-then-fall is the shape of narrowing for faces and speech sounds in infants \autocite{pascalis2002species,werker1984cross,maurer2014perceptual}. In hearing the representation narrows (separability $d'$ falls from 5.19 to 1.33, against 5.15 in the exposed control), but the network is never surprised by the unfamiliar change, so whether ``narrowing'' is found depends on whether discrimination is measured like looking time or like habituation. Both signatures held with a predictive-coding network \autocite{rao1999predictive} and a transformer.

\paragraph{Each sense needs a different model.} In the cross-sense study, the smallest network that reached the 12-month level differed by sense: about 6,000 core parameters for hearing, about 24,000 for touch, while vision was still improving at 97,000. More capacity did not rescue newborn vision (hidden-object tracking $R^2$ $-2.67$ at 1,700 parameters and $-2.09$ at 97,000), so a young sense is made young by its input, not by a small network. Every sense needed some memory of the recent past; beyond that, which architecture worked best depended on the sense. A blurry-first sequence of visual input improved tracking of hidden objects in vision (0.51 against 0.33) while changing nothing in hearing or touch (3 seeds).

\paragraph{The first sense crowds out a later one.} When two senses share one read-out and one comes online first, the later sense's encoder is under-trained: with hearing first, as in infants, the vision encoder reached 0.60 against 0.84 for a vision-only network, even with equal visual experience, while training each sense separately and combining them afterwards gave the best encoders (0.84 and 0.84). A similar result appeared in a three-sense model (vision predicted worst on its own when it came online last: $R^2$ $-0.18 \pm 0.14$ against $0.43 \pm 0.05$ when first, 5 seeds). This is consistent with critical periods for multisensory integration in networks \autocite{kleinman2023critical} and with the proposal that staggered sensory onset limits competition between senses \autocite{turkewitz1982limitations}. A hint that the infant order helps fill in vision from hearing and touch did not survive five seeds ($0.10 \pm 0.15$).

\subsection{The Baby Senses Lab: the same filters as an interactive simulator}

\begin{figure}[tbp]
\centering
\includegraphics[width=\textwidth]{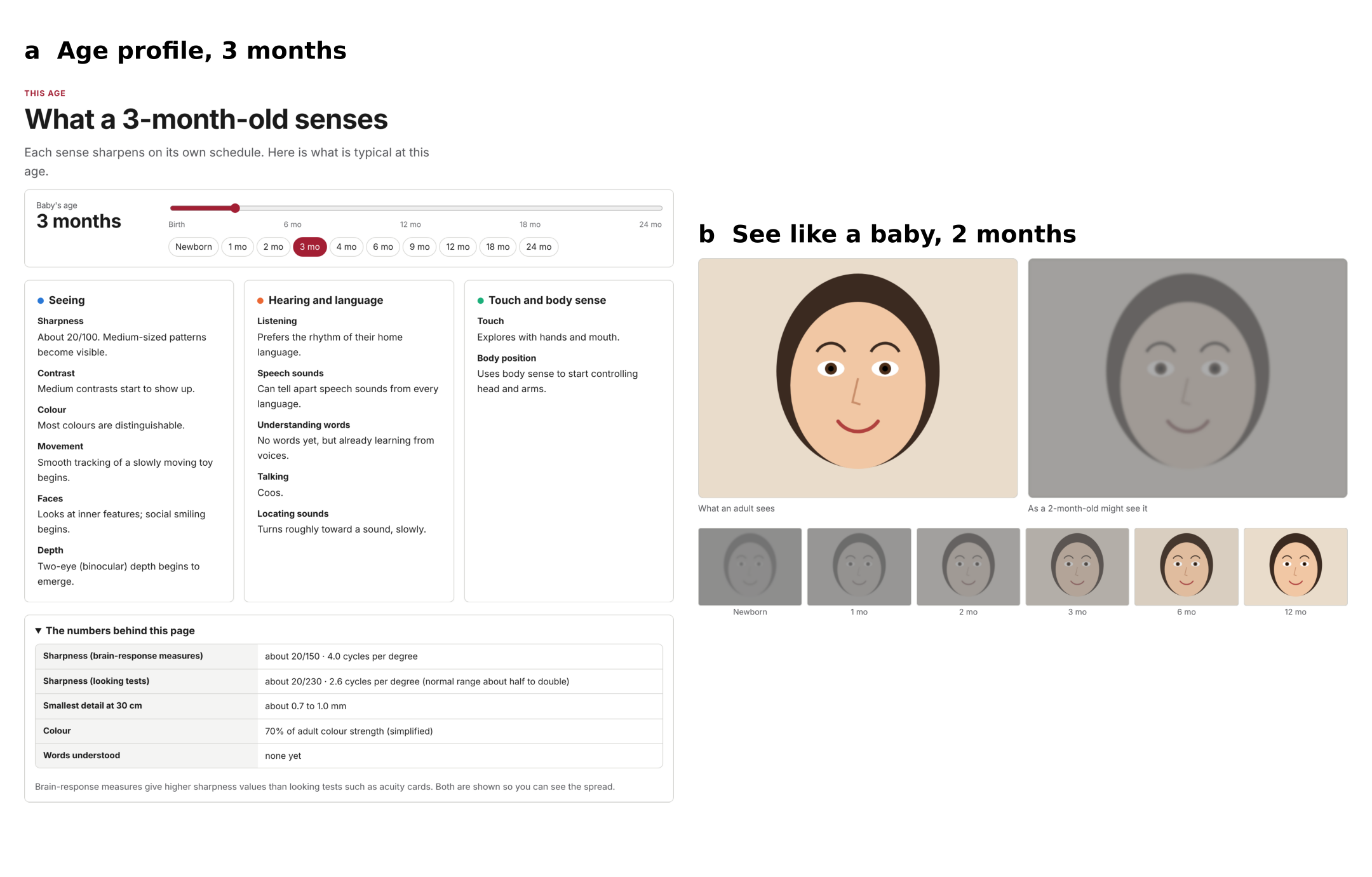}
\caption{\textbf{The Baby Senses Lab.} \textbf{a}, Age profile at 3 months: plain-language descriptions of seeing, hearing, and touch and body sense (thinking, social and movement are in a separate section), and the numbers behind them, opened from the section below the descriptions, including both acuity measures and the smallest detail resolvable at 30 cm. \textbf{b}, ``See like a baby'': an image shown as an adult sees it and as a 2-month-old might at 30 cm, using the front end of Fig.~\ref{fig:frontends}a--c, and the same image across the first year (newborn to 12 months). Users can load their own photos and videos. Screenshots of the live site.}
\label{fig:app}
\end{figure}

The age tables and front ends are useful outside the models, and we packaged them as a free website, the Baby Senses Lab (\url{https://pikulsomesh.github.io/developmental-world-models/}), that runs entirely in the browser (Fig.~\ref{fig:app}). Nothing a user loads is uploaded.

\paragraph{What it offers.} On a first visit the website asks whether the visitor is a parent or other curious person or a researcher, and shows a view for each; the choice is remembered and can be changed on any page. The family view is described here; the researcher view summarises the findings, methods, papers and code. A user chooses an age from birth to 24 months. (i) An \emph{age profile} describes what babies of that age typically see, hear and feel; sensory domains are shown first, and thinking, social and movement domains are kept separate, in a section that opens on request, and labelled as such. (ii) \emph{See like a baby} shows the user's own photo or video, or a built-in drawing, blurred, reduced in contrast and desaturated by the amounts in Fig.~\ref{fig:frontends}a--c for a chosen viewing distance, with a choice between the evoked-potential and behavioural acuity tables. (iii) \emph{Hear like a baby} plays a built-in, recorded or uploaded sound through the hearing front end of Fig.~\ref{fig:frontends}d--f and shows its spectrum. (iv) An \emph{activity library} lists 27 everyday activities with the abilities each needs, and a word checker gives approximate typical ages of comprehension for about 100 early words, based on large parent-report norms and laboratory tests \autocite{frank2021variability,bergelson2012}. (v) \emph{Try together} offers five short observation activities adapted from infant research methods: preference for a face-like pattern over a scrambled one \autocite{johnson1991newborns}, smooth tracking of a moving toy \autocite{vonhofsten1997development}, preferential looking at stripes against grey \autocite{teller1997first}, turning to a sound from one side, and looking time at an expected or unexpected outcome \autocite{baillargeon1987object}. An observer records what the baby does; the session is summarised next to typical findings and can be downloaded as a table.

\paragraph{What it is for.} The website gives a quick, concrete answer to ``what can a baby of this age sense?'' that is grounded in the same numbers as the models. For a parent or teacher, it makes, for example, the 20--30 cm viewing distance or the value of high-contrast patterns in the first months tangible. For a student, it turns tables of thresholds into images and sounds. For a researcher, it is a way to inspect the input that an age-tuned model receives, to compare measurement methods side by side, and to pilot observation protocols.

\paragraph{What it is not.} The website does not screen or diagnose, and every page says so. Displays are not calibrated instruments, one session with one baby is a noisy sample, and the age tables summarise populations rather than individuals. The views simulate the limits of the sense organs and early sensory processing, not a baby's experience: babies move their eyes differently, focus poorly at some distances in the first weeks, and their brains fill in a great deal.

\section{Discussion}

\paragraph{Age-tuned models as a quick simulation of sensory development.} The approach taken here is to write each sense's immaturity as a fixed filter indexed by age, and place a learning network behind it. The filter makes the simulation fast and transparent, because anyone can see and hear what reaches the network at a given age; the network adds what a filter cannot, namely what is extracted from that input and how experience shapes it. Used together, they separate three questions that are entangled in infant data: what information reaches a sense, what is learned from it, and how the order and timing of experience change what is learned.

\paragraph{What the simulations reproduce, and why.} Three results should be kept apart. The order of sensory maturation (touch and hearing ahead of vision) is reproduced by construction, because it is in the front-end tables; it is not evidence that networks develop like infants, and a cruder measure based on available information does not recover it at all. The slow early rise of vision beyond what the input allows is a property of learning, and it suggests that early visual limitations in infants may also be partly limitations of extraction, a hypothesis that can be tested by comparing behavioural and evoked-potential acuity with the same stimuli \autocite{dobson1978visual}. Critical periods, perceptual narrowing and competition between senses are history effects that appear in one continuous run of a plain prediction network. These last results are the most interesting computationally, because nothing in the networks is switched on or off at a particular age.

\paragraph{Perception, not cognition.} This paper is restricted to sensory perception. The same networks that represent where a hidden toy is are not surprised when it disappears, except when its sound stops too \autocite{mohapatra2026learn}: sensory prediction does not by itself produce the expectations about objects that infants show \autocite{baillargeon1987object}. A simulation of what a baby senses should therefore not be read as a simulation of what a baby knows, and the website keeps the two apart.

\paragraph{Why sense-specific models.} Each sense has its own clock, its own input and its own critical periods \autocite{hensch2005critical,werker2015critical}, and in our simulations the minimum model size, the best architecture and the effect of a gentle start all differed by sense. Training a model per sense first and combining them later gave the best encoders. Sense-specific, age-tuned models are thus not only easier to interpret; they avoid the competition that arises when senses that mature at different times share one learner.

\begin{figure}[tbp]
\centering
\includegraphics[width=0.55\textwidth]{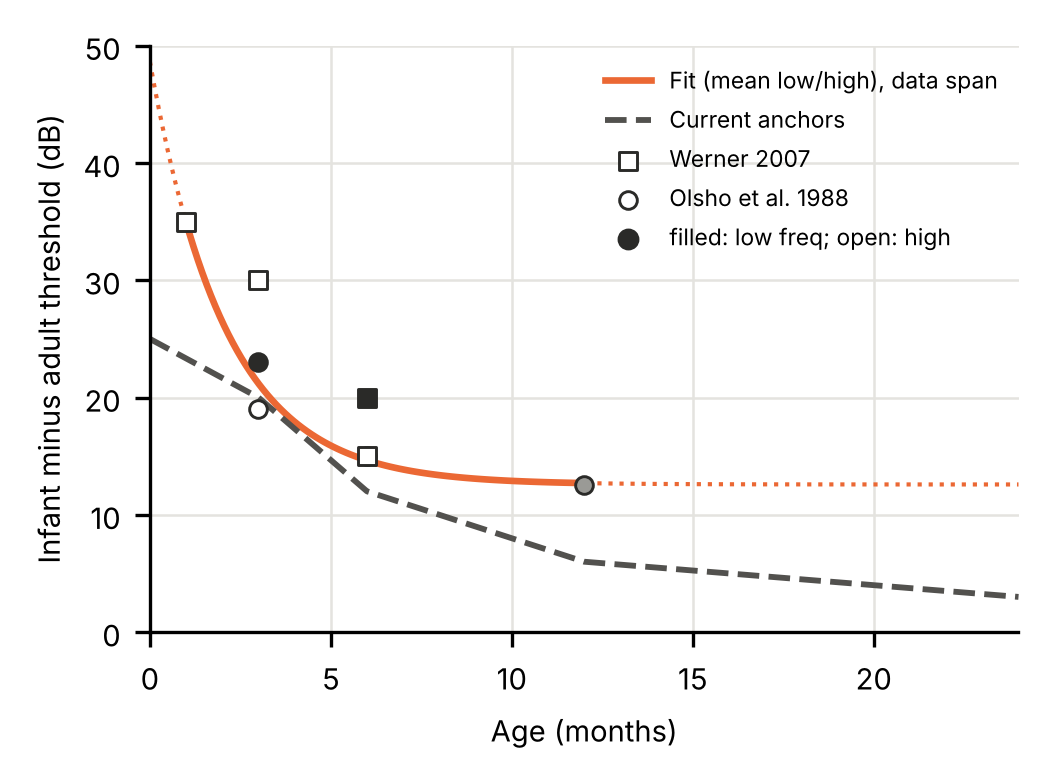}
\caption{\textbf{Hearing front end compared with infant threshold data.} Points: infant-minus-adult pure-tone thresholds from group studies \autocite{olsho1988pure,werner1990puretone,werner2007issues}; filled markers low frequencies (250--500~Hz), open markers high frequencies (4--8~kHz); grey, the pooled 6- and 12-month value. Review values that differ from the primary study \autocite{werner2007issues} are shown but not fitted. Orange line: exponential fit to the mean of low and high frequencies over the range of the data (dotted beyond it). Dashed line: the attenuation anchors used by the models and the website (Fig.~\ref{fig:frontends}d). There are no data at birth or in the second year.}
\label{fig:hearfit}
\end{figure}

\paragraph{The hearing anchors are close to threshold data in early infancy but too mild later.} To check the hearing front end in the same way as vision, we compared its extra attenuation with published infant-minus-adult pure-tone thresholds (Fig.~\ref{fig:hearfit}). At 1 month infants are about 35~dB behind adults at 4~kHz \autocite{werner1990puretone,werner2007issues}; at 3 months about 23~dB at 500~Hz and 19~dB at 4~kHz; and at 6 and 12 months about 10--15~dB, with the larger gap at low frequencies \autocite{olsho1988pure}. An exponential fit to the mean of low and high frequencies, $36.1\,e^{-a/2.10} + 12.6$~dB for age $a$ in months (only 4~kHz at 1 month), agrees with the current anchors within 3~dB at 3 and 6 months but is 12~dB higher at 1 month and 7~dB higher at 12 months. The data therefore support a steeper fall over the first months and a residual gap of about 13~dB that persists through the first year, where the anchors fall towards adult values. They do not support a large high-frequency cut: at 3 months the deficit at 4~kHz is smaller than at 500~Hz, by 6 months the two are similar, and only at 8~kHz at 3 months is the high-frequency deficit larger \autocite{olsho1988pure}. The trained hearing networks used the original anchors; because their learned read-out tracked the ceiling at every stage (Fig.~\ref{fig:ceiling}), a stronger early attenuation would be expected to lower both together rather than to open a gap between them, but this was not re-run. Touch has no comparable threshold series across the first year in the sources we could verify.

\begin{figure}[tbp]
\centering
\includegraphics[width=\textwidth]{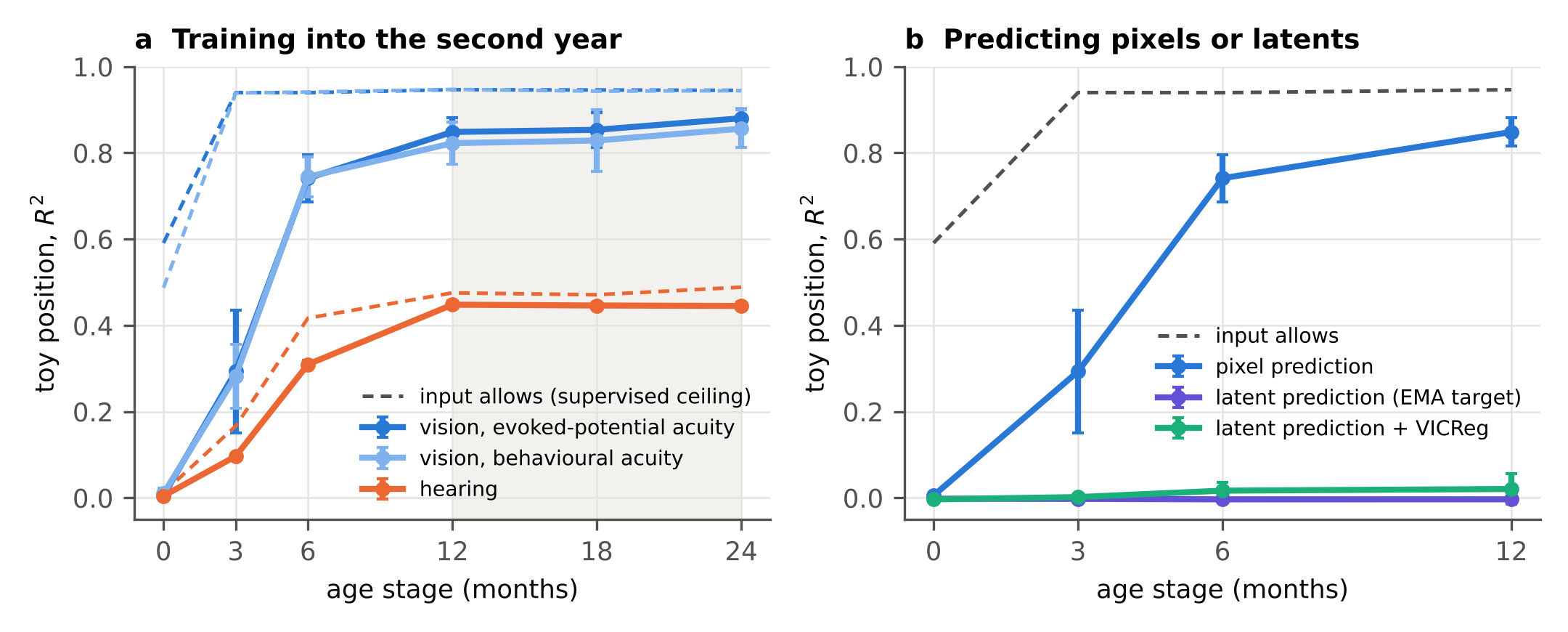}
\caption{\textbf{Training beyond 12 months, and predicting latents instead of pixels.} \textbf{a}, Toy position decoded from the main-study networks (own-age linear probe; mean $\pm$ 1 s.d., 5 seeds) continued to 18 and 24 months (shaded), for vision with the evoked-potential (dark blue) and behavioural (light blue) acuity tables and for hearing (orange); dashed lines, supervised information ceiling for the same input. \textbf{b}, Vision on the 0--12 month schedule: pixel prediction (the main-study objective) against latent prediction with a slowly updated target encoder, with and without variance--covariance regularisation; dashed line, information ceiling. New runs for this paper.}
\label{fig:beyond}
\end{figure}

\paragraph{A second year of training adds little in this world.} We continued the main-study vision and hearing networks (the same full-size networks, with input quality following age) from 12 to 18 and 24 months, 5 seeds each (Fig.~\ref{fig:beyond}a). Vision was run with two acuity tables: the evoked-potential table of the main study, which is constant after 12 months, and the behavioural table of the website, which keeps rising to 10.5 cycles/degree at 24 months. The 12-month values with the evoked-potential table are those of the five-seed main study ($0.85 \pm 0.04$), which the follow-up reproduces because it uses the same seeds. Position read-out then barely changed: $0.85 \pm 0.05$ at 18 months and $0.88 \pm 0.02$ at 24 months with the evoked-potential table, and $0.82 \pm 0.06$, $0.83 \pm 0.08$ and $0.86 \pm 0.05$ at 12, 18 and 24 months with the behavioural table. Hearing stayed at $0.45 \pm 0.01$, close to its ceiling (0.49). The behavioural table's continuing improvement in acuity added no information that a network could use, because the toy is large compared with the remaining blur: its ceiling was already 0.94 at 3 months, as for the evoked-potential table. What did keep improving was toy identity in vision ($0.64 \pm 0.06$ at 12 months to $0.70 \pm 0.09$ at 24 months with the evoked-potential table; $0.57 \pm 0.07$ to $0.61 \pm 0.06$ with the behavioural table), with no change in input for the evoked-potential table, so the gain comes from further experience rather than better input. The second year of sensory development is mostly fine refinement (Fig.~\ref{fig:timeline}), and a world with one large toy cannot test it; tasks that need fine detail, such as small or distant objects, would be needed.

\paragraph{In this small world, the latent-prediction variants we tried did not recover the toy.} Objectives that predict the next state in a learned latent space rather than in pixels are a leading alternative for world models \autocite{assran2023self,garrido2025intuitive}. We trained the main-study vision network on the same 0--12 month schedule without its decoder, predicting the latent code that a slowly updated copy of its own encoder gives for the next frame, with no gradient through that copy \autocite{grill2020bootstrap} (Fig.~\ref{fig:beyond}b; 5 seeds). This version collapsed: within a few hundred steps every frame mapped to almost the same code (per-dimension s.d.\ 0.006 or less, against 0.12 for pixel prediction), and position could not be read out at any stage ($R^2 \le 0.00$). Adding variance and covariance penalties that prevent collapse \autocite{bardes2022vicreg} kept the code spread out, over about 15 effective dimensions against 7 for pixel prediction, but it still carried almost nothing about the toy ($0.02 \pm 0.04$ at 12 months, best seed 0.09, against $0.85 \pm 0.04$ for pixel prediction), and identity stayed at chance (0.25). Our reading, not tested here, is that the pixel network is pushed towards the toy by its reconstruction loss, which up-weights what moves, while the latent objective has no such push and spends its code on the larger, easier parts of the scene. Masked prediction over many patches and much more data, which the published latent-prediction models use, may change this. This is one small configuration (32$\times$32 frames, one toy, a small network and a 0--12 month schedule), so it does not show that latent prediction is unsuitable for early learning or for world models in general; finding an architecture or masking strategy that survives newborn-level blur is future work.

\paragraph{Limitations.} All results come from small networks in very simple synthetic environments (low-resolution frames, one toy), evaluated mostly with linear read-outs and with 3--7 seeds per condition; effect sizes are estimates. The vision tables follow published infant data, but the two families of acuity measurement disagree, and we use one in the models. The hearing anchors match threshold data at 3 and 6 months but are too mild at 1 month and from 12 months on (Fig.~\ref{fig:hearfit}), the touch and body-position tables are coarse anchors that respect the direction of known changes, and taste, smell and the vestibular sense are not modelled. The models' ages are stages of one training run and are not calibrated to hours of infant experience, and the networks were trained beyond 12 months only in the follow-up runs of Fig.~\ref{fig:beyond}a, in one synthetic world, so the second year rests mainly on the literature synthesis and the website's tables. Generative world models that render new views of persistent 3D scenes, such as World Labs' RTFM, are described only at a high level, with no published architecture \autocite{worldlabs2025rtfm}, and our fixed-camera world has no viewpoint for them to change, so they were not compared. The website's images are rendered on uncalibrated screens, its colour change is a simplification of infant chromatic sensitivity, and its observation sessions have not been validated against laboratory measures.

\paragraph{Next steps.} Three steps would turn this into a validated tool. First, train the age-tuned models on head-camera recordings from children \autocite{sullivan2021saycam,long2024babyview} and, for touch and body position, in an embodied infant simulator \autocite{mattern2024mimo}, and compare their read-outs with infant behaviour using the same paradigms (preferential looking, habituation, violation of expectation). Second, compare predictive-coding architectures \autocite{rao1999predictive,lotter2017deep} and latent-prediction objectives with masking, on richer scenes and more data, with the pixel-prediction networks used here. Third, run a small validation study of the website's observation sessions with calibrated displays, and test its usefulness with parents and educators.

\section{Materials and Methods}

\subsection{Experimental design}

The paper combines a literature synthesis with three modelling studies from one project and an application built on the same parameters. The synthesis (Fig.~\ref{fig:timeline}, Table~\ref{tab:senses}) collects sensory milestones reported in group studies, stating the measurement method where methods disagree. The modelling studies are summarised in Table~\ref{tab:studies}; their full methods and code are available with the companion paper \autocite{mohapatra2026learn}. All data are synthetic.

\begin{table}[t]
\caption{\textbf{Modelling studies summarised in this paper.}}
\label{tab:studies}
\centering
\footnotesize
\begin{tabularx}{\textwidth}{@{}p{2.6cm}LLp{1.5cm}@{}}
\toprule
Study & Senses and input & Networks and objective & Seeds \\
\midrule
Main study (Fig.~\ref{fig:ceiling}) & Vision (32$\times$32 colour), binaural hearing (2$\times$32 bands), touch (8$\times$8 patch), body position; one toy, a moving hand and an occluding screen & Encoder, residual dynamics over a memory window and decoder; reconstruct and predict next frame; four age stages in one run; supervised information ceiling per age & 5 \\
Cross-sense study (Fig.~\ref{fig:experience}a, b, d, e) & Vision (12$\times$12), hearing (24 bands), touch (16 taxels) rendered from one latent world & Recurrent, transformer, predictive-coding and latent-prediction networks at $\approx$20k core parameters; next-step prediction & 3 (5 for onset order) \\
Age-modelling study (Fig.~\ref{fig:experience}c) & Vision (shape-versus-texture gratings); vision plus audio for onset order & Compact convolutional networks, fixed or growing; classification; blur windows of 3--9 months & 3--7 \\
Follow-up runs (Fig.~\ref{fig:beyond}) & As the main study; vision and hearing & Main-study networks continued to 18 and 24 months; and latent prediction against pixel prediction, 0--12 months & 3 \\
\bottomrule
\end{tabularx}
\end{table}

\subsection{Vision front end}

This is the vision front end defined in the companion paper \autocite{mohapatra2026learn}, restated because the website implements the same equations. Three age-dependent quantities, all interpolated linearly between the ages in Fig.~\ref{fig:frontends}, set it: acuity $c(a)$ (cycles per degree), contrast sensitivity $\mathrm{CS}(a)$ and colour saturation $\gamma(a)$. The image $o$ is first blurred by a Gaussian $G_\sigma$ whose modulation transfer $\exp(-2\pi^2\sigma^2 f^2)$ falls to 0.1 at $f = c(a)$:
\begin{equation}\label{eq:blur}
\sigma(a) = \frac{\sqrt{\ln 10 / 2}}{\pi\, c(a)}\ \text{degrees},\qquad x = G_{\sigma(a)} * o .
\end{equation}
In the website $\sigma$ is converted to pixels with the displayed image width $w$, the real width of the scene $W$ and the viewing distance $D$, $\text{pixels per degree} = w / \big(2\arctan(W/2D)\cdot 180/\pi\big)$. Contrast is pulled toward the image mean $\bar x$ with a gain that is logarithmic in contrast sensitivity, because contrast sensitivity is a threshold rather than a gain:
\begin{equation}\label{eq:contrast}
x \leftarrow \bar x + \kappa(a)\,(x - \bar x),\qquad \kappa(a) = \mathrm{clip}\!\left(\tfrac12 \log_{10}\mathrm{CS}(a),\ 0.15,\ 1\right),
\end{equation}
and colour is faded toward luminance $\ell = 0.299x_R + 0.587x_G + 0.114x_B$:
\begin{equation}\label{eq:colour}
\tilde o = \mathrm{clip}_{[0,1]}\big(\ell + \gamma(a)\,(x - \ell)\big).
\end{equation}
The models' front end uses the same three steps on 32$\times$32 images, with the blur width set from the acuity ratio to adult (30 cycles per degree) and an assumed field of view \autocite{mohapatra2026learn}. The smallest resolvable detail reported by the website is one half-cycle at the acuity limit, $D\tan\!\big(1/(2c(a))\big)$.

\subsection{Hearing, touch and body-position front ends}

In the models, each ear's spectrum is smoothed across frequency, high frequencies are rolled off, the interaural difference is scaled toward the mean of the two ears by a factor $\rho(a)$, and Gaussian noise is added. In the website, real stereo audio passes through the same operations implemented with the Web Audio API: a mid/side mix that scales the left-right difference by $\rho(a)$ (left output $\tfrac{1+\rho}{2}L + \tfrac{1-\rho}{2}R$, and symmetrically), a high-shelf filter at 3 kHz with gain $-g(a)$ dB, an overall attenuation of $A(a)$ dB, and a soft low-pass noise floor. Touch receives receptive-field blur and noise on the tactile patch; body position receives additive noise.

\subsection{Networks, read-outs and information ceiling}

The main study's sense networks encode the filtered observation, predict the next latent state from a window of past states and the infant's own movement, and decode back to the input; they are trained to reconstruct the current input and predict the next, through four successive stages of one run that we call 0, 3, 6 and 12 months \autocite{mohapatra2026learn}. The main-study runs, own-age read-outs and information ceilings used in Fig.~\ref{fig:ceiling} are shared with the companion paper \autocite{mohapatra2026learn}, which uses them to analyse learning; here they are used only to compare what each sense's input holds with what is extracted from it. A linear probe (ridge regression) reads the toy's position or identity from the network's state; $R^2 = 1 - \sum\lVert y-\hat y\rVert^2/\sum\lVert y - \bar y\rVert^2$. ``Own-age'' probes are fitted and tested on input of the stage's own age. The information ceiling at each age is the held-out score of a network of the same encoder family trained directly, with supervision, to recover the toy's position and identity from one filtered frame. To measure surprise, we take the network's error in predicting the next input and average it over the frames around the event; for each pair of matched events (one possible, one impossible), $d_z$ divides the average gap between the two by the spread of that gap across pairs. Narrowing uses one continuous run in which only two of four identities ever appear, with surprise at identity swaps and the separability $d'$ of the two identities' representations tracked during training.

\subsection{Follow-up runs}

The follow-up runs use the main study's full-size sense networks, data generator, training steps (2,000 per stage for vision, 1,500 for hearing) and own-age probes, with new code that does not change the main study's. For the second year, the run continues at 18 and 24 months; vision uses either the evoked-potential acuity table, constant after 12 months, or the website's behavioural table (0.8 cycles/degree at birth, 7.5 at 12 months, 9.0 at 18 and 10.5 at 24), with contrast and colour unchanged, and the hearing tables are constant after 12 months. The information ceiling is recomputed for each table and age. For latent prediction, the decoder is removed and the dynamics module predicts the next frame's code from a target encoder whose weights are an exponential moving average (rate 0.99) of the encoder's, with no gradient through the target \autocite{grill2020bootstrap}; the regularised version adds variance and covariance penalties on the codes with the published weights (25, 25 and 1) \autocite{bardes2022vicreg}. Collapse is measured by the mean per-dimension s.d.\ of the codes of held-out frames and by their effective dimension, $(\sum_i\lambda_i)^2/\sum_i\lambda_i^2$ over the eigenvalues $\lambda_i$ of their covariance.

\subsection{The Baby Senses Lab}

The website is static HTML, CSS and JavaScript served from GitHub Pages. Images and videos are filtered with the CSS \texttt{filter} property (blur, contrast, saturate) computed from Eqs.~\ref{eq:blur}--\ref{eq:colour}; sound with the Web Audio API. The age profile uses the tables of Fig.~\ref{fig:frontends} for 0--12 months and developmental surveillance milestones for 15--24 months \autocite{zubler2022evidence}. Each ability $k$ used by the activity library has a typical age $m_k$ and width $w_k$, and its level at age $a$ is $\ell_k(a) = 1/\big(1+\exp(-(a-m_k)/w_k)\big)$; an activity's readiness is the mean over its required abilities of $\min(1, \ell_k(a)/n_k)$, where $n_k$ is the level the activity needs. Observation sessions run full-screen, sized from the screen width and viewing distance the observer enters, and record the observer's key presses with timestamps; sessions are stored only in the browser and can be downloaded as CSV or JSON.

\subsection{Statistical analysis}

Values are mean $\pm$ 1 s.d. over independent training seeds; the number of seeds is given with each result. No new statistical tests are reported here; planned comparisons for the main study are reported in the companion paper \autocite{mohapatra2026learn}.

\subsection{Animal and human studies}

The study uses only synthetic data and published group results; no human or animal participants were involved. The website's observation activities are offered for informal use and were not used to collect data for this paper.

\clearpage
During the preparation of this work, the author used an AI assistant (Claude, Anthropic) to help write and run analysis code, build the website, draw figures and draft text. The author reviewed and edited the content and takes full responsibility for the content of the publication.

All code, results, figures and the website source are at \url{https://github.com/pikulsomesh/developmental-world-models}. The website is at \url{https://pikulsomesh.github.io/developmental-world-models/}. All data are synthetic and are regenerated by the code from the listed random seeds.

S. Mohapatra conceived the project, designed the studies and the website, and approved the final text.

This work received no specific funding and was carried out independently.

The author declares that there is no conflict of interest regarding the publication of this article.

\printbibliography

\end{document}